%% file: root.tex
\documentclass[letterpaper, 10 pt, conference]{style/ieeeconf}  

\IEEEoverridecommandlockouts                              

\title{\LARGE \bf Single-Shot Learning of Stable Dynamical Systems \\ for Long-Horizon Manipulation Tasks}

\author{Alexandre St-Aubin$^{1}$, Amin Abyaneh$^{2}$, and Hsiu-Chin Lin$^{1,2}$ 
\thanks{$^{1}$School of Computer Science, McGill University, Montreal, Canada }
\thanks{$^{2}$Department of Electrical and Computer Engineering, McGill University, Montreal, Canada }
\thanks{{\tt\small alexandre.st-aubin2@mail.mcgill.ca}}
\thanks{This work is sponsored by NSERC Discovery Grant, FRQNT Research Support for New Academics, FRQNT Doctoral Training Scholarships, and McGill Science Undergraduate Research Awards.}
}

\input{style/symbol}

\begin{document}

\maketitle
\thispagestyle{empty}
\pagestyle{empty}

\begin{abstract}
\input{tex/abstract}
\end{abstract}

\section{Introduction}
\input{tex/introduction}

\section{Background}
\input{tex/background}

\section{Methods}
\input{tex/method}

\section{Experiments}
\input{tex/experiment}

\section{Conclusion}
\input{tex/conclusion}




\clearpage
\addcontentsline{toc}{section}{References}
\bibliographystyle{style/IEEEtran}
\bibliography{bib/imitation,bib/long-horizon}

\end{document}

%% file: style/symbol.tex
\usepackage{graphics} 
\usepackage{epsfig} 
\usepackage{mathptmx} 
\usepackage{times} 
\usepackage{amsmath} 
\usepackage{amssymb}  
\usepackage{amsfonts}
\usepackage[acronym]{glossaries}
\usepackage{caption}
\usepackage{subcaption}
\usepackage{stfloats}
\usepackage{xcolor}
\usepackage{cite}
\usepackage{mathtools}
\usepackage{hyperref}
\usepackage{multirow}
\usepackage{color}
\usepackage[normalem]{ulem}
\usepackage{algorithm2e}
\usepackage{algpseudocode}
\usepackage{booktabs}

\usepackage[english]{babel}
\usepackage{array}

\newcommand{\xpos}{\mathbf{x}}
\newcommand{\xvel}{\dot{\xpos}}

\newcommand{\controller}{\mathcal{C}}
\newcommand{\trajectory}{\tau}
\newcommand{\R}{\mathbb{R}}

\newcommand{\xposGoal}{\mathbf{g}}
\newcommand{\Demo}{\mathcal{D}}

\newcommand{\dimTraj}{\mathit{N}}
\newcommand{\indexTraj}{n}
\newcommand{\dimTask}{\mathit{K}}
\newcommand{\indexTask}{k}

\newcommand{\keyframe}{\mathit{h}}
\newcommand{\keyframeK}{\keyframe^\indexTask}
\newcommand{\subgoal}{\mathbf{g}}
\newcommand{\subgoalK}{\mathbf{g}^\indexTask}
\newcommand{\DemoK}{\Demo^{\indexTask}}
\newcommand{\Waypoints}{\mathbf{W}}
\newcommand{\WaypointsK}{\mathbf{W}^\indexTask}
\newcommand{\waypoint}{\mathbf{w}}
\newcommand{\AWEthreshold}{\eta}

\newcommand{\threshold}{\epsilon}

\newtheorem{remark}{Remark}[subsection]
\newtheorem{proposition}{Proposition}[subsection]

\newcommand{\BCSeg}{{\bf BC+Ours}}
\newcommand{\Ours}{{\bf Ours}}
\newcommand{\SNDS}{\bf SNDS}
\newcommand{\BC}{\bf BC}

\newcommand{\dsStableParam}{\theta}
\newcommand{\dsStableSym}{\pi_{\dsStableParam}}

\newcommand{\dsStable}{\dsStableSym(\xpos)}
\newcommand{\lpfFunction}{v}

\newcommand{\lossSNDSSegment}{\mathcal{L}}

\newcommand{\thresholdK}{\threshold^\indexTask}

\newacronym{STL}{STL}{Standard Triangle Language/Standard Tessellation Language}
\newacronym{URDF}{URDF}{Unified Robot Description Format}
\newacronym{AABB}{AABB}{Axis-Aligned Bounding Boxes}
\newacronym{OBB}{OBB}{Oriented Bounding Boxes}
\newacronym{GJK}{GJK}{Gilbert Johnson Keerthi}
\newacronym{RGJK}{RGJK}{Recursive Gilbert Johnson Keerthi}
\newacronym{EPA}{EPA}{Expanding Polytope Algorithm}
\newacronym{SVD}{SVD}{Singular Value Decomposition}
\newacronym{ANN}{ANN}{Artificial Neural Network}
\newacronym{RL}{RL}{Reinforcement Learning}
\newacronym{MDP}{MDP}{Markov Decision Process}
\newacronym{PPO}{PPO}{Proximal Policy Optimization}
\newacronym{IL}{IL}{Imitation Learning}
\newacronym{BC}{BC}{Behavioural Cloning}
\newacronym{DAGGER}{DAGGER}{Dataset Aggregation}
\newacronym{GAIL}{GAIL}{Generative Adversarial Imitation Learning}
\newacronym{IRL}{IRL}{Inverse Reinforcement Learning}
\newacronym{GCL}{GCL}{Guided Cost Learning}
\newacronym{GAN}{GAN}{Generative Adversarial Networks}
\newacronym{SEDS}{SEDS}{Stable Estimator of Dynamic Systems}
\newacronym{MSE}{MSE}{mean squared error}

%% file: tex/abstract.tex

\noindent
Mastering complex sequential tasks continues to pose a significant challenge in robotics. While there has been progress in learning long-horizon manipulation tasks, most existing approaches lack rigorous mathematical guarantees for ensuring reliable and successful execution. In this paper, we extend previous work on learning long-horizon tasks and stable policies, focusing on improving task success rates while reducing the amount of training data needed. Our approach introduces a novel method that (1) segments long-horizon demonstrations into discrete steps defined by waypoints and subgoals, and (2) learns globally stable dynamical system policies to guide the robot to each subgoal, even in the face of sensory noise and random disturbances. We validate our approach through both simulation and real-world experiments, demonstrating effective transfer from simulation to physical robotic platforms.

%% file: tex/introduction.tex

\noindent
Learning to perform complex sequential tasks continues to be a fundamental challenge in robotics~\cite{nairhierarchical, lee2021ikea, lynch2020learning, kou2024kisa}. 
Many routine tasks, such as assembly and decluttering, require sequential decision-making and coordinated interaction between the robot and objects to perform a series of motion primitives.

\emph{Imitation learning} works through these sequential tasks by learning to emulate a set of expert demonstrations~\cite{morrison2018closing,ding2019goal,jang2022bc}. 
However, most imitation learning methods are designed to learn a single task and tend to be unreliable when learning from complex, long-horizon expert demonstrations~\cite{eysenbachc,nairhierarchical,lynch2020learning}. Notably, safety and stability guarantees are frequently disregarded in favor of achieving higher performance in stochastic environments~\cite{dawson2022safe}.   


Recent efforts to tackle long-horizon tasks focus on goal-conditioned imitation learning~\cite{zhang2024universal,mezghani2023learninggoalconditionedpoliciesoffline}.
Since long-horizon robotics tasks often involve multiple implicit subtasks or skills, a more viable direction refines the learning process by decomposing long-horizon and sequential tasks into a series of \emph{subtasks} capable of reconstructing the original expert trajectories~\cite{pertsch2020long, kou2024kisa, shi2023waypoint, zhang2024universal}.


Despite this progress, most of these methods cannot provide rigorous mathematical guarantees to generate reliable and successful outcomes.
Therefore, deploying policies learned from long-horizon manipulation tasks in simulation to real robot systems raises a lot of safety concerns. 
Additionally, most methods require an impractically large number of demonstrations, even for simple manipulation tasks such as pick-and-place. 


The core challenge resides in the fact that long-horizon tasks require accomplishing a cascade of multiple \emph{subgoals} before achieving the main objective, without the explicit knowledge about the subgoals.
This brings about three significant challenges. 
First, as the horizon lengthens, the robot must handle increasing levels of uncertainty, an underlying cause of \emph{compounding error}. 
Secondly, the robot must infer the subgoals from the demonstrations, introducing an additional layer of uncertainty to the problem.
Third, if the robot fails to reach one subgoal, the whole task is jeopardized. 


\begin{figure}[t]
      \centering
      \includegraphics[width=0.48\textwidth]{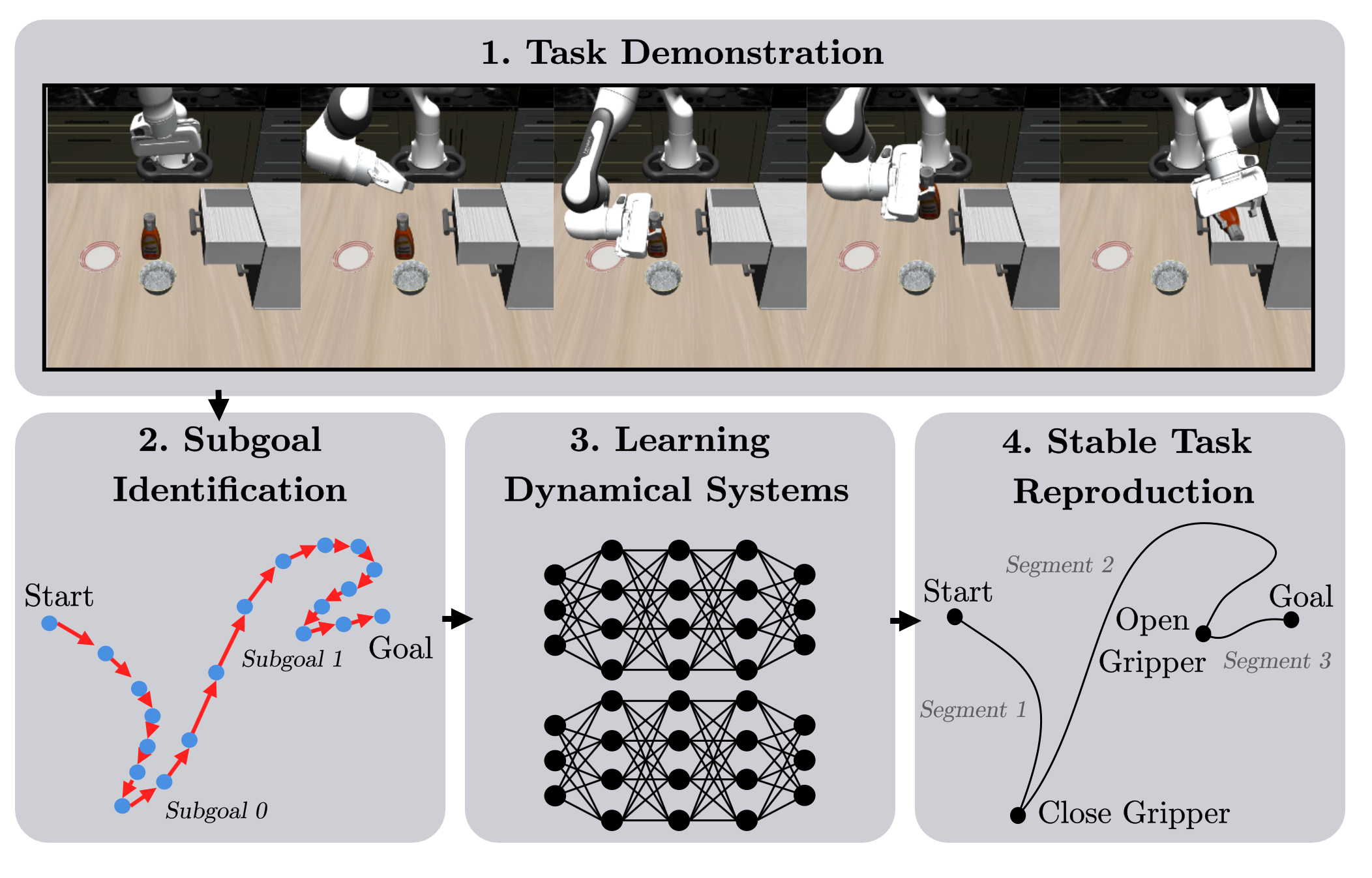}
      \caption{Overview of our approach: Long-horizon demonstrations (1) are first segmented into subgoals (2). Low-level stable dynamical policies are then learned to robustly reach each subgoal, even in the presence of perturbations (3). Finally, a high-level policy orchestrates a cascade of these stable policies for each segment, replicating the long-horizon expert demonstrations (4).}
      \label{fig:introduction}
   \end{figure}

When learning from long-horizon demonstrations, it is crucial to ensure that each intermediary step is executed safely. 
Prior work on \emph{stable} dynamical policies has been broadly utilized to imitate expert behavior while remaining resilient to perturbations~\cite{khansari2011learning,rana2020euclideanizing,coulombe2023generating,Abyaneh2024}. The stability property ensures that all trajectories induced by dynamical policies converge to an equilibrium state. 
However, these methods are designed to learn from a single primitive. Even the most expressive dynamical policies struggle to learn long-horizon tasks~\cite{rana2020euclideanizing, Abyaneh2024}, as ensuring global stability becomes increasingly difficult over extended time horizons.  
  
In this paper, we build upon prior work on learning long-horizon manipulation tasks and learning stable policies, aiming to enhance task success rates while minimizing the required training data.
We focus on the problem of the movement planning of the robot, without considering visual feedback and/or understanding human intention. 

We propose a novel approach that (1) splits a long-horizon demonstration into a set of segments, characterized by waypoints and a subgoal and (2) learns a set of globally stable dynamical system policies that guide the robot to each subgoal despite sensory noise and stochastic perturbations (see Fig.~\ref{fig:introduction}).
The proposed method is validated both in simulation and on robotic hardware, with a direct transfer from simulation to real-world implementation. 
The main contributions include:
    \begin{itemize}
        \item We extend the previous work on learning dynamical systems~\cite{Abyaneh2024} to problems of long-horizon tasks where the robot reaches each subgoal with rigorous theoretical guarantees.        
        \item We show that our learned policies can be deployed directly onto a real-world system with a {\em single demonstration}.
\end{itemize}
Our code is available at \href{https://github.com/Alestaubin/stable-imitation-policy-with-waypoints}{github.com/Alestaubin/stable-imitation-policy-with-waypoints}.


%% file: tex/background.tex
\label{sec:background}
\noindent
In this section, we provide a formal problem statement, and a brief background on learning stable dynamical policies and waypoint extraction methods later used in this paper.

\subsection{Problem Definition}

\noindent
Given a demonstration of a long-horizon manipulation task 
$\Demo=\{(\xpos_1,\;\xvel_1), \;(\xpos_2, \;\xvel_2), \;\dots, \;(\xpos_\dimTraj, \;\xvel_\dimTraj) \}$, where $(\xpos_\indexTraj, \; \xvel_\indexTraj)$ denote the position and the velocity at the $n^{th}$ step in the demonstration, and $\dimTraj$ is the task's horizon length.
The variables $\xpos \in\R^d, \; \xvel = \frac{\partial \xpos}{\partial t} \in\R^d$ unambiguously define the d-dimensional position and velocity of a robotic system. For example, $\xpos$ could be the joint angles or end-effector pose. 

We assume that the demonstration $\Demo$ can be decomposed into a set of $\dimTask$ \emph{sub-demos}, $\{\Demo^1, \Demo^2, \dots, \Demo^\dimTask\}$, such that $\DemoK$ represents the $k^{th}$ subtask with its corresponding subgoal $\xposGoal^\indexTask$. 
Our goal is to provide the robot with the velocity $\xvel$ given the current position $\xpos$ at each time step, to reproduce the expert demonstration {\em without} prior knowledge about the dimensionality of the task $\dimTask$ and the subgoal $\subgoalK $. 

\subsection{Stable Neural Dynamical Systems (SNDS)}
\label{sec:background_snds}
\noindent Assuming $\Demo$ represents the expert trajectory, Stable Neural Dynamical Systems (SNDS)~\cite{Abyaneh2024} efficiently learns a stable dynamical policy,
\begin{align}
\label{eq:snds_policy}
\xvel = \dsStable, \quad \dsStableSym: \R^d \xrightarrow{} \R^d.
\end{align}
Modeled by an ordinary differential equation, $\dsStable$ is optimized such that for any initial state, $\xpos_0 \in \Demo$, the forward Euler method~\footnote{This holds for any method used to solve an initial value problem, with the forward Euler method selected for simplicity.} generates a sequence,
\begin{align}
\label{eq:euler_trajectory}
    \trajectory_{\xpos_{0}}^{\dsStableSym}  = \{\xpos_0, \xpos_1, \ldots, \xpos_M \;\vert\; \xpos_{m+1} = \xpos_m + \dsStableSym(\xpos_m) \Delta t\}, 
\end{align}
which accurately replicates the corresponding trajectory in $\Demo$ for sufficiently large $M$. On top of that, SNDS employs Lyapunov theory~\cite{devaney2021introduction} to ensure global stability~\cite{manek2019learning, Abyaneh2024}, meaning that the sequence $\trajectory_{\xpos_{0}}^{\dsStableSym}$ converges to a predefined equilibrium, $\xposGoal$, for any arbitrary initial condition, $\xpos_0 \notin \Demo$. 

A dynamical policy exhibits global stability if there exists a positive-definite function $\lpfFunction: \R^d \xrightarrow{} \R$, known as a Lyapunov candidate, such that $\dot{\lpfFunction}(\xpos) < 0$ for all $\xpos \neq \xposGoal$ and $\dot{\lpfFunction}(\xpos) = 0$. SNDS learns both the policy, $\dsStableSym$, and the Lyapunov candidate, $\lpfFunction$, by minimizing the following hybrid loss, $\lossSNDSSegment(\dsStableSym, \Demo)$, on expert data: 

\begin{gather}
\label{eq:snds_hybrid_loss}
     \gamma \mathop{\mathbb{E}}_{\xpos, \xvel \in \Demo } \Big[\big(\dsStableSym(\xpos) - \xvel)^2\Big] \; +
    (1 - \gamma) \mathop{\mathbb{E}}_{\xpos_i, \tau_{\xpos_i}^{\Demo} \in \Demo}  \Big[\big(\trajectory_{\xpos_i}^{\dsStableSym} - \trajectory_{\xpos_i}^{\Demo})^2\Big], 
\end{gather}

\noindent 
in both position and velocity spaces. In Eq.~\ref{eq:snds_hybrid_loss}, $\gamma \in [0,1]$ controls the trade-off between the position and velocity components in the loss function. The terms $\trajectory_{\xpos_i}^{\Demo}$ and $\trajectory_{\xpos_i}^{\dsStableSym}$ represent partial trajectories from the demonstration data $\Demo$ and those generated by $\dsStableSym$, respectively.

\subsection{Automatic Waypoint Extraction}
\label{sec:background-awe}
\noindent
Automatic Waypoint Extraction (AWE) aims to find waypoints that can reproduce a demonstration. 
Specifically, it aims to select a set of waypoints $\Waypoints$ and reconstruct a trajectory by interpolating between every consecutive pairs of waypoints. 
AWE is formulated as an optimization problem
\begin{equation}
    \begin{aligned}
        \min_\Waypoints & || \Waypoints|| \\
                & s. t. ~ \mathcal{L} ( f(\Waypoints) , \Demo) \leq \AWEthreshold
    \end{aligned}
    \label{equ:waypoint}
\end{equation}
where $f(\Waypoints)$ is a function that takes a set of waypoints and reconstructs a trajectory, $\mathcal{L}$ is the loss function, and $\AWEthreshold$ is a threshold of maximum loss.

The parameter $\AWEthreshold$ allows the user to decide the precision of the reconstructed motion.
A lower $\AWEthreshold$ indicates that the user prefers high-precision imitation, in which the method will automatically select more waypoints to fit the demonstration $\Demo$.
In contrast, a higher $\AWEthreshold$ will yield a smaller set of waypoints from Equ.~\ref{equ:waypoint}, and the reconstructed trajectory will be a rougher approximation of $\Demo$.

\subsection{Discussion}
\noindent 
The prior work on data-driven methods for learning stable policies (such as the one in Sec.~\ref{sec:background_snds} was designed for solving a single task. 
In our work, we will adapt the same network architecture for each subtask of a long-horizon problem (see Sec.~\ref{method:DS}).

AWE, discussed in Sec.~\ref{sec:background-awe} automatically selects waypoints given demonstrations.
The hyper-parameter $\AWEthreshold$ in Equ.~\ref{equ:waypoint} allows the user to make a tradeoff between accuracy and smoothness.
If a demonstration contains a discontinuous movement, AWE typically marks it as a waypoint. 
However, such motion can be the intention of the demonstration or the results of {\em} noise present in the data. 
In our work, we adapt AWE to select waypoints within a segment of a long horizon manipulation task (Sec.~\ref{method:segmentation}). 
This allows smooth movement within a segment without sacrificing the precision for achieving the subtask.

%% file: tex/method.tex

\noindent
We aim to learn long-horizon manipulation tasks in a one-shot manner using a hierarchical approach. 
Our approach leverages~\cite{garrett2021integrated}, breaking down the policy into high-level decision-making and low-level motion planning. At the high level, we define the task as a series of subgoals (Sec~\ref{method:segmentation}). Next, a unique and stable dynamical policy is learned for each segment based on the corresponding sub-demo (Sec~\ref{method:DS}). Lastly, at execution time, we reconstruct the trajectory by selecting the appropriate policy to perform each part of the task (Sec.~\ref{method:reconstruction}).

\subsection{Subgoal Identification and Waypoint Selection}
\label{method:segmentation}

\noindent
Our first step is to identify key states in the trajectory where major stages of the overall task take place, thereby breaking down complex trajectories into more manageable segments for learning. 
We opt for a straightforward method, defining a subgoal as the activation of the gripper. 
Our insight is that in most household tasks, meaningful subgoals typically involve the hand or gripper either grasping or releasing an object. 
By defining these actions as subgoals, we can divide the demonstrations into sub-demos, where each segment can be easily described by a single dynamical policy.

Formally, we define untrimmed demonstrations as 
$\Demo=\{(\xpos_1,\;\xvel_1), \;(\xpos_2, \;\xvel_2), \;\dots, \;(\xpos_\dimTraj, \;\xvel_\dimTraj) \}$,
we need to find the key frames that divide the demonstration $\Demo$ into $\dimTask$ sub-demos, such that  $\DemoK$ resembles a simple motion conditioned on a subgoal $\subgoalK$.

We perform a forward pass in the trajectory to find indices $\keyframe^1, \keyframe^2, \dots, \keyframe^\dimTask$ such that $\xpos_{\keyframeK}$ denotes the $\indexTask^{th}$ occasions where the gripper opens or closes.
Based on the selected indices, we divided the trajectory $\Demo$ into $\dimTask$ segments, and  define the subgoal for each segment as $\subgoalK = \xpos_{\keyframeK}$.

\begin{figure}[t]
     \centering
     \begin{subfigure}[b]{0.32\linewidth}
         \centering
         \includegraphics[width=\textwidth]{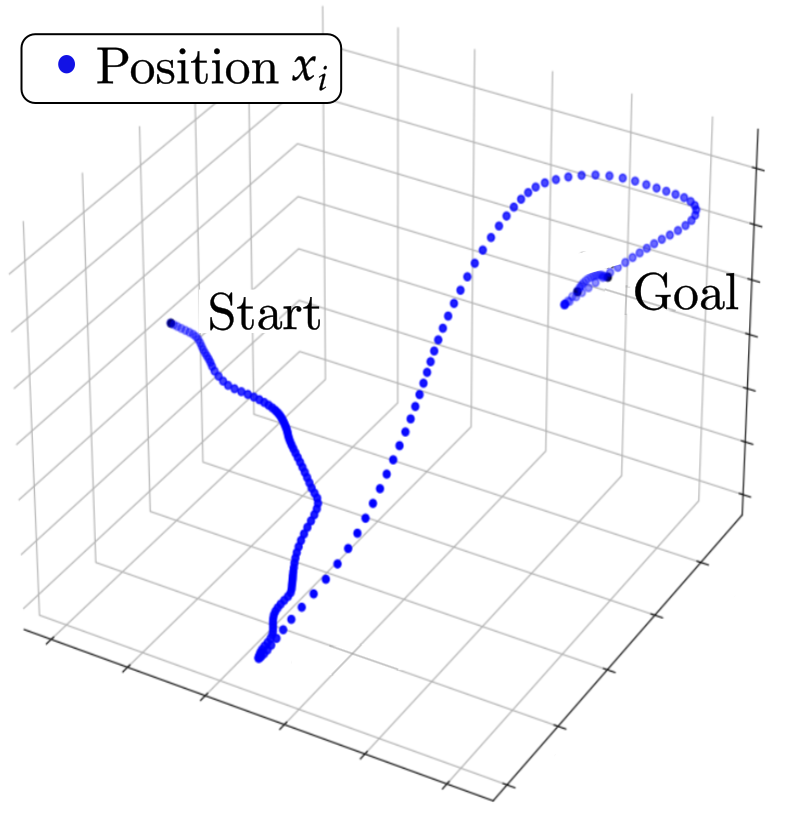}
         \caption{Expert trajectory}
         \label{fig:data}
     \end{subfigure}
     \hfill
     \begin{subfigure}[b]{0.32\linewidth}
         \centering
         \includegraphics[width=\textwidth]{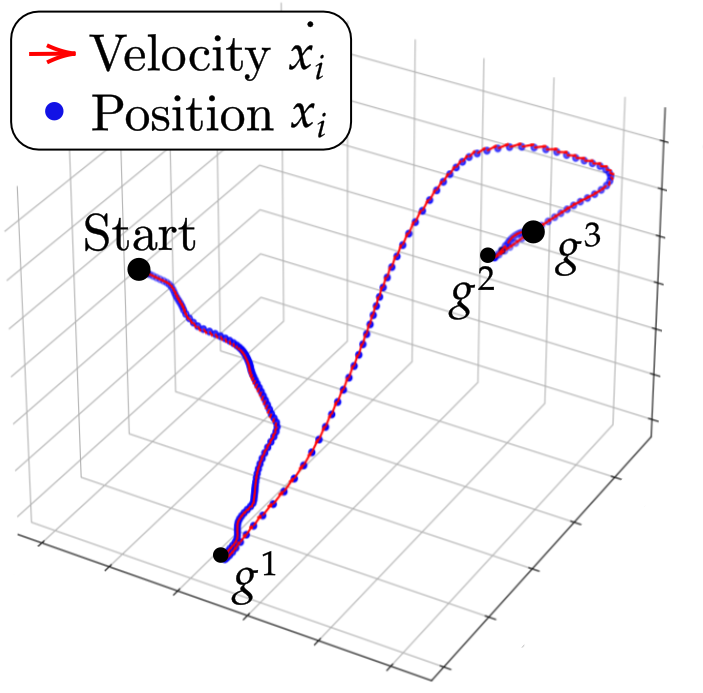}
         \caption{Selected subgoals}
         \label{fig:segmentation}
     \end{subfigure}
     \hfill
     \begin{subfigure}[b]{0.33\linewidth}
         \centering
         \includegraphics[width=\textwidth]{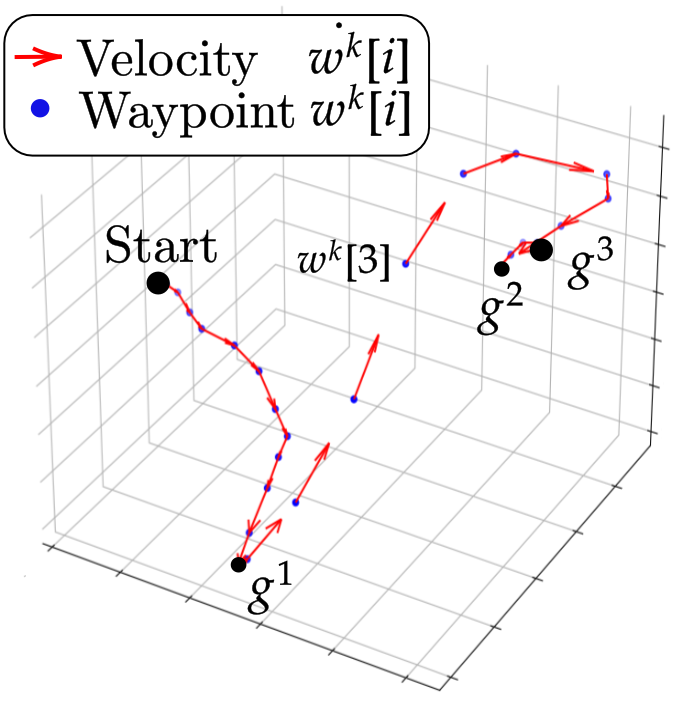}
         \caption{AWE waypoints}
         \label{fig:waypoint-outcome}
     \end{subfigure}   
    \caption{An example of (a) a single expert demonstration in the robot's task space, (b) three subgoals selected, and (c) outcomes of the automatic waypoint selections in each segment.}
    \label{fig:waypoint-comparison}
\end{figure}

An example can be seen in Fig.~\ref{fig:waypoint-comparison}. The original demonstration (blue) contains a discontinuous motion (Fig.~\ref{fig:data}), and our method labels the transition points with 3 subgoals $\subgoal^1,\subgoal^2,\subgoal^3$ (Fig.~\ref{fig:segmentation}).

From each segment $\Demo^\indexTask$, we filter out the noise and reduce the data complexity by finding waypoints that approximate $\Demo^\indexTask$, thereby simplifying the task even further.
We leverage AWE (see Sec.~\ref{sec:background-awe}) to automatically extract waypoints from data (using Equ.~\ref{equ:waypoint}).
For each segment, a set of waypoints is selected  $\WaypointsK = \{\waypoint^\indexTask_0,\; \waypoint^\indexTask_1,\;..., \;\subgoalK \} $ and the last waypoint is simply the subgoal of the segment. 

AWE selects waypoints sparingly when the trajectory is straight and more densely when the trajectory is complex, providing more data where necessary and less where it’s not. 
By adjusting the trajectory reconstruction loss threshold $\AWEthreshold$, the smoothness of the trajectory can be controlled. 

Fig.~\ref{fig:waypoint-outcome} is an example of waypoint extraction.
Note that only a small amount of data is retained since this is sufficient to reconstruct the original demonstration through interpolations between waypoints.

\begin{remark}
In contrast to the previous work, the waypoint extraction is performed solely within a segment.
Our insight is that the most important requirement of manipulation tasks lies in achieving the subgoal, while precise imitation may not be essential.
We can filter the noise or sacrifice the accuracy of imitation by reconstructing the demonstration with the sampled waypoints, but the subgoal cannot be approximated.    
\end{remark}

\subsection{Learning Dynamical Systems}
\label{method:DS}
\begin{figure}[t]
    \centering      
    \includegraphics[width=0.41\textwidth]{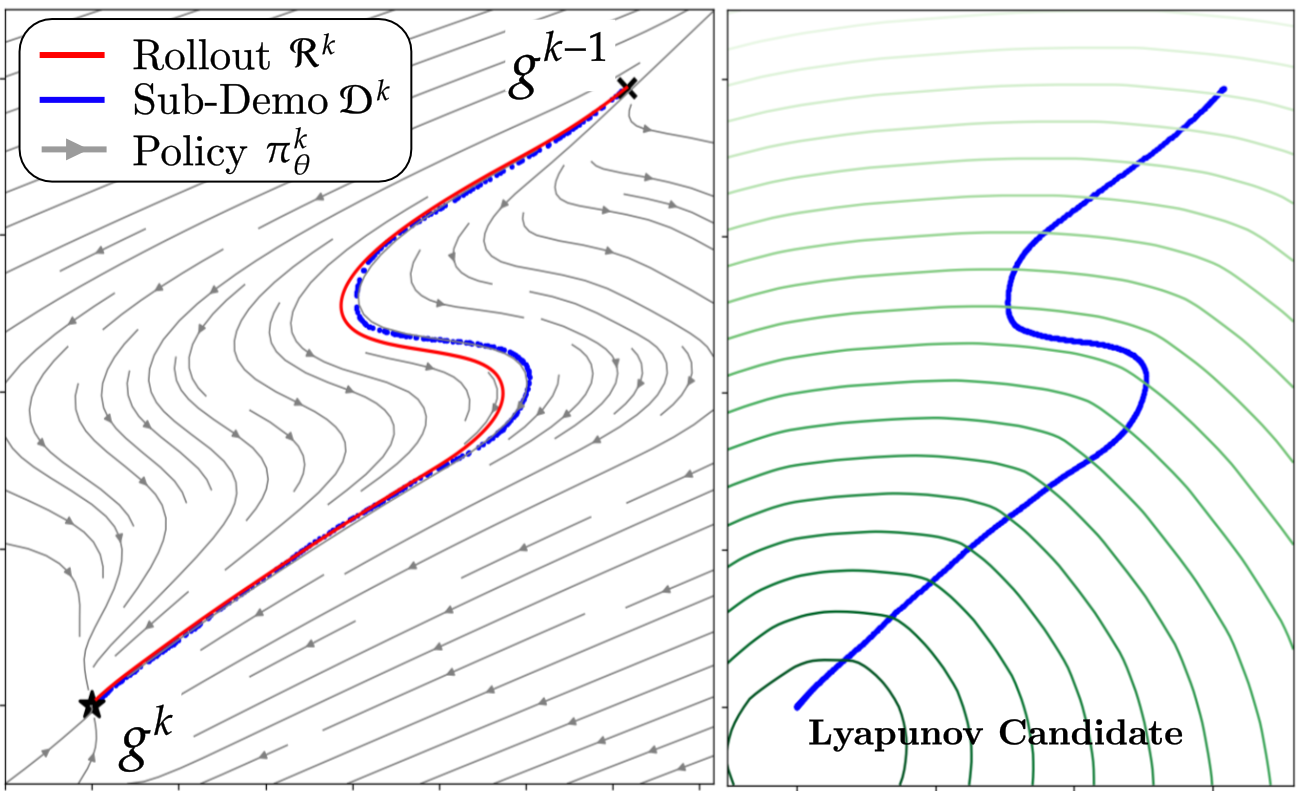}
    \caption{Stable dynamical policy rollout by an optimized SNDS model (left), and its Lyapunov candidate (right). The learned Lyapunov candidate ensures the induced trajectories always move toward the lowest energy point, regardless of the initial state or perturbations.}
    \label{fig:streamplot_snds}
\end{figure}

\noindent 
The second step is to learn a set of $\dimTask$ models that can reproduce the motion from the segmented trajectories derived from Sec.~\ref{method:segmentation}.
For each $\indexTask^{th}$ segment, we train an SNDS policy, $\tilde{\xvel} = \dsStableSym^\indexTask(\xpos)$ with the following objective, with the loss $\lossSNDSSegment$ defined in Eq.~\ref{eq:snds_hybrid_loss}.

\begin{equation} \label{eq:snds_loss_segment}
    \dsStableParam^*_k \triangleq \arg\min_{\dsStableParam \in \R^{N_\dsStableParam}}
    \lossSNDSSegment(\dsStableSym^\indexTask, \;\DemoK)
\end{equation}

\noindent The training process is conducted exclusively on the data from the $\indexTask^{th}$ segment, denoted as $\Demo^k$, with the subgoal $\subgoalK$ set as the stable equilibrium. Since $\dsStableSym$ is formulated using standard automatic differentiation tools, the optimization problem can be efficiently solved to determine the optimal parameter, $\dsStableParam^*$. 

An example is illustrated in as illustrated in Fig.~\ref{fig:streamplot_snds}. 
The left figure shows the segmented demonstration $\DemoK$ (blue) and the policy rollout (red) from the dynamical system $\dsStableSym^\indexTask$. 
The grey arrows represent the vector field produced by the dynamical system. 
In regions where demonstrations are absent, the motion is mostly determined by the Lyapunov candidate function (right) which enforces movement toward the subgoal.

\begin{remark}
As explained in Sec~\ref{sec:background_snds}, the representation of SNDS is specifically designed to enforce global stability. 
Consequently, each learned policy, $\dsStableSym^\indexTask$, generates velocities $\tilde{\xvel}$ based on the current state to imitate expert data within the segment, ensuring that all trajectories converge to the subgoal $\subgoalK$.
For this reason, even in the presence of external disturbances or noisy inputs, the dynamical system can still bring the robot to the subgoal.
\end{remark}

\subsection{Stable Task Reproduction}
\label{method:reconstruction}
\noindent
Having learned a unique dynamical system for each segment, what remains is to define a high-level controller $\controller$ to imitate the task by returning desired velocities at each state during execution.

The high-level controller $\controller$ takes as input the set of subgoals (from Sec.~\ref{method:segmentation}) and learned dynamical systems (from Sec.~\ref{method:DS}).
At each time step, the high-level controller evaluates the current state $\xpos$ and determines which subgoal should be the target and whether the current subgoal was achieved, based on a distance threshold $\thresholdK$. 
The parameter $\thresholdK$ can be adjusted depending on the nature of the task.

\begin{figure}[t]
    \centering      
    \includegraphics[width=0.47\textwidth]{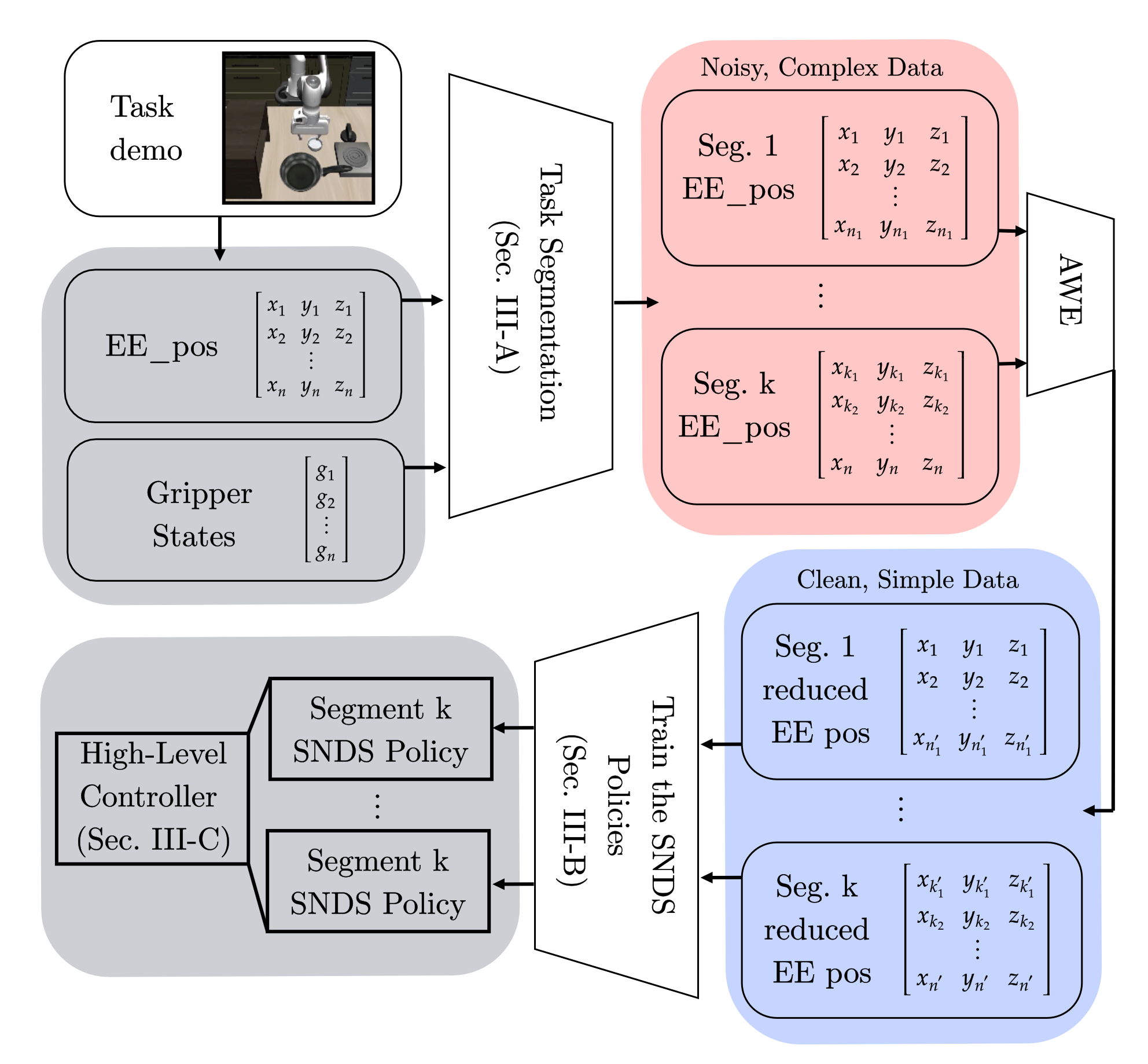}
    \caption{Our framework on learning stable policies for long-horizon manipulation tasks}
    \label{fig:method-scheme}
\end{figure}

Then, the high-level controller applies policy $\tilde{\xvel} = \dsStableSym(\xpos, \subgoalK)$ and executes the predicted velocity $\tilde{\xvel}$ during the execution of segment $k$ of the trajectory. Specifically, 
\begin{align} \label{eq:highlevel_control}
\tilde{\xvel} = 
\begin{cases}
  \dsStableSym^{\indexTask}(\xpos), & \|\xpos - \subgoalK\| > \thresholdK \\
  \dsStableSym^{\indexTask+1}(\xpos), & \|\xpos - \subgoalK\| \leq \thresholdK \; \land \; \indexTask < \dimTask - 1 \\
  \dsStableSym^{\indexTask}(\xpos), & \indexTask = \dimTask - 1 
\end{cases}
\end{align}

\begin{proposition}
The high-level policy outlined in Eq.~\ref{eq:highlevel_control} is globally stable at the last subgoal, $\subgoal^\dimTask$.
\end{proposition}
\noindent \textit{Proof.} The proof is intuitive and follows from the global stability of each low-level dynamical policy. Formally, according to the Lyapunov global stability theorem, 
$$ \forall \xpos_0^{\indexTask} \in \DemoK, \;\; \lim_{t \rightarrow \infty} \xpos_t 
 = \subgoalK, \;\;\text{if}\;\; \xpos_{t+1} = \xpos_t + \Delta t \;\dsStableSym^{\indexTask}(\xpos), $$

\noindent   for sufficiently small $\Delta t$. Then, every $\subgoalK \in \DemoK$ of $\dsStableSym^{\indexTask}(\xpos)$ can be viewed as the initial condition of $\dsStableSym^{\indexTask + 1}(\xpos)$: $\subgoal^{\indexTask + 1}$. Hence, the core definition of global stability can be written seen as a cascade, yielding: $\forall \xpos_0^{\indexTask} \in \Demo, \;\; \lim_{t \rightarrow \infty} \xpos_t 
 = \subgoal^{\dimTask}, \;\;\text{when}\;\; \xpos_{t+1} = \xpos_t + \Delta t \;\dsStableSym(\xpos) $. \hfill $\square$

 \vspace{5pt}



%
%

A summary of our proposed method is illustrated in Fig.~\ref{fig:method-scheme}.
This architecture ensures resilience against noise and external perturbations. 
This allows the system to quickly return to the original path and avoid collisions in cluttered environments. 

%% file: tex/experiment.tex

\begin{figure}[t]
     \centering
     \begin{subfigure}[t]{0.32\linewidth}
         \centering
         \includegraphics[width=\textwidth]{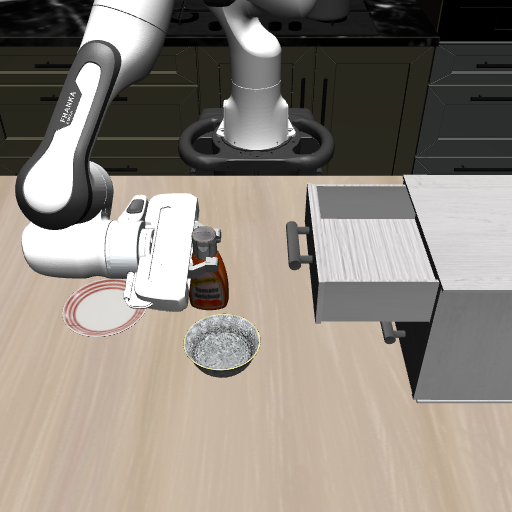}
         \caption{Ketchup}
         \label{fig:taskA}
     \end{subfigure}
     \hfill
     \begin{subfigure}[t]{0.32\linewidth}
         \centering
         \includegraphics[width=\textwidth]{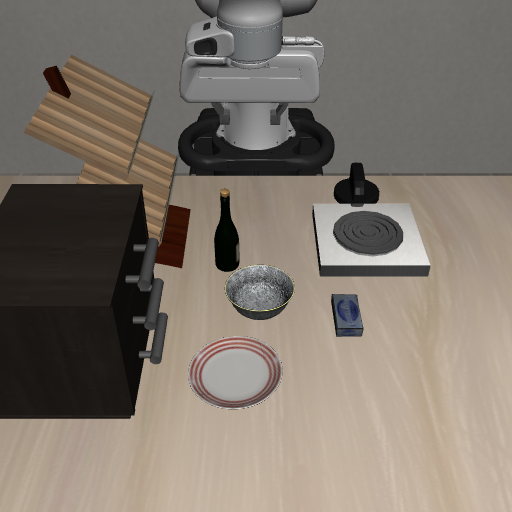}
         \caption{Wine \& Cheese}
         \label{fig:taskB}
     \end{subfigure}
     \hfill
     \begin{subfigure}[t]{0.32\linewidth}
         \centering
         \includegraphics[width=\textwidth]{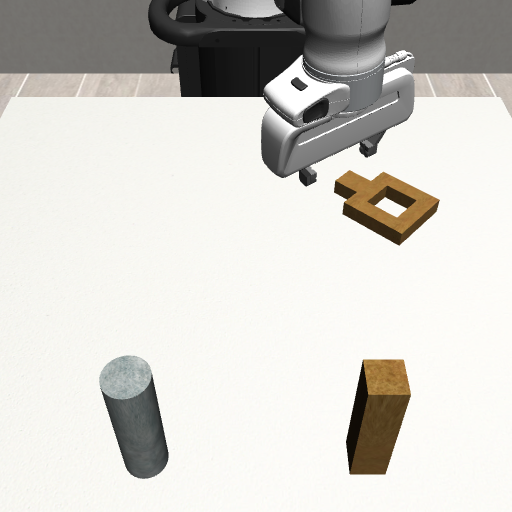}
         \caption{Square}
         \label{fig:taskC}
     \end{subfigure}   
     
     \vspace{10pt}  
     
     \small  
     \begin{tabular}{@{}>{\centering\arraybackslash}p{1.5cm}
                        >{\centering\arraybackslash}p{2.0cm}
                        >{\centering\arraybackslash}p{0.8cm}
                        >{\centering\arraybackslash}p{1.0cm}
                        >{\centering\arraybackslash}p{1.5cm}@{}}
     \toprule
     \textbf{Task}   & \textbf{Dataset} & \textbf{Demo \#}   & \textbf{Demo Length} & \textbf{Waypoints} \\ 
     \midrule
     Ketchup & LIBERO-90       & 1 & 230 & 25 \\ 
     Square  & Robomimic       & 0 & 127 & 20 \\ 
     Wine    & LIBERO-Goal     & 4 & 158 & 24\\ 
     Bowl    & LIBERO-90       & 0 & 92  & 18\\ 
     Cheese  & LIBERO-Goal     & 1 & 92  & 14 \\ 
     \bottomrule
     \end{tabular}
     
     \caption{Examples of tasks in Robosuite (top) and an overview of the task demonstrations (bottom).}
     \label{fig:exp-robosuite-tasks}
\end{figure}

\noindent Our experiments aim to demonstrate the following questions:
\begin{enumerate}
    \item Can we improve the overall success rate by enforcing the success of each subtask?
    \item Can we learn a long-horizon manipulation task from a single demonstration?
\end{enumerate}

\subsection{Evaluation Criteria}
\noindent To evaluate the performance, we consider both the success rate on the success rate of completing (1) the sub-task within each segment and (2) the whole task. 

\subsection{Baselines}
\noindent 
We compare our work against the following baselines. 
\begin{enumerate}
    \item {\bf BC}: standard behavioral cloning
    \item {\bf SNDS}: original Stable Neural Dynamical Systems~\cite{Abyaneh2024}
    \item {\bf \BCSeg}: To evaluate our proposed work on trajectory segmentation with waypoint selections (Sec.~\ref{method:segmentation}), we also compare the performance of using standard behavior cloning on each segment data. 
\end{enumerate}

\noindent
Lastly, we use $\Ours$ to denote the model trained with our method described in both Sec.~\ref{method:segmentation}) and Sec.~\ref{method:DS}


\begin{figure}[t]
    \centering
    \begin{subfigure}[b]{\linewidth}
        \centering
        \includegraphics[width=\textwidth]{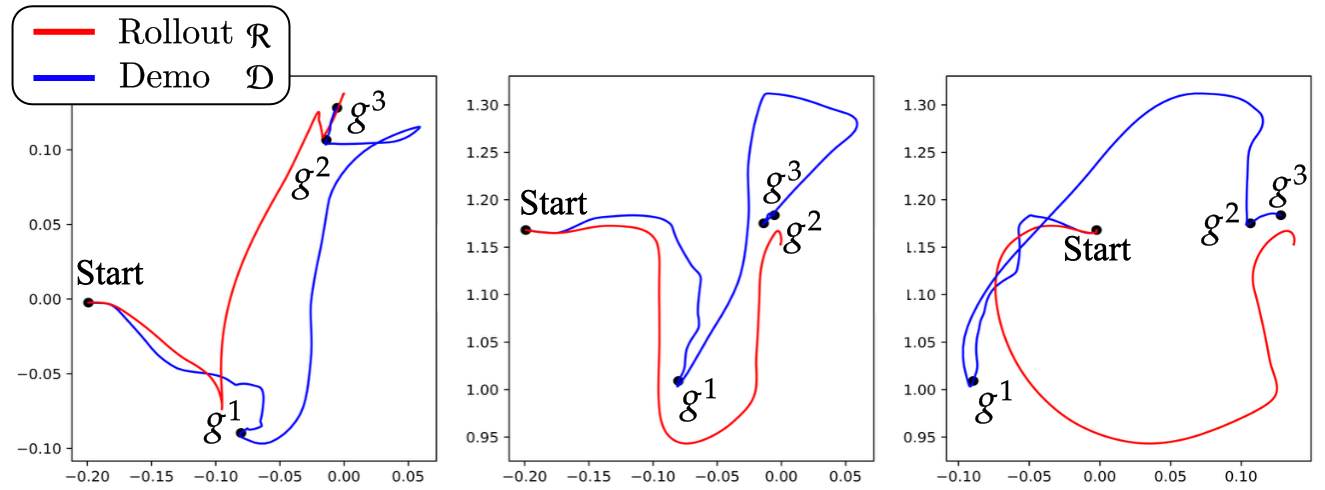}            
        \caption{\bf BC}
        \label{fig:nn}
    \end{subfigure}
    \begin{subfigure}[b]{\linewidth}
        \centering
        \includegraphics[width=\textwidth]{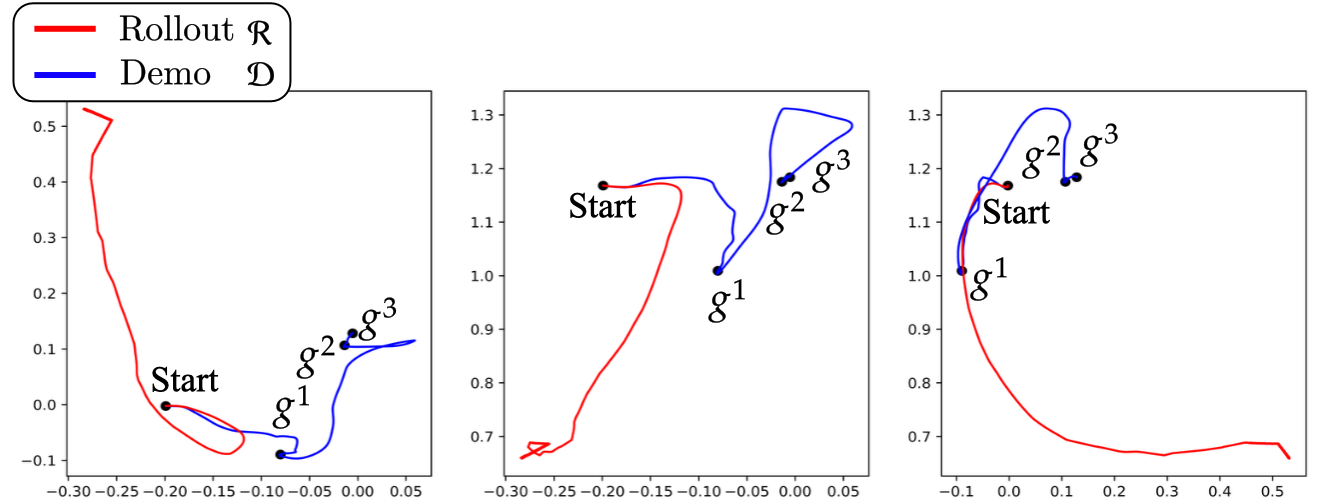}
        \caption{\bf SNDS}
        \label{fig:snds}
    \end{subfigure}
    \begin{subfigure}[b]{\linewidth}
        \centering
        \includegraphics[width=\textwidth]{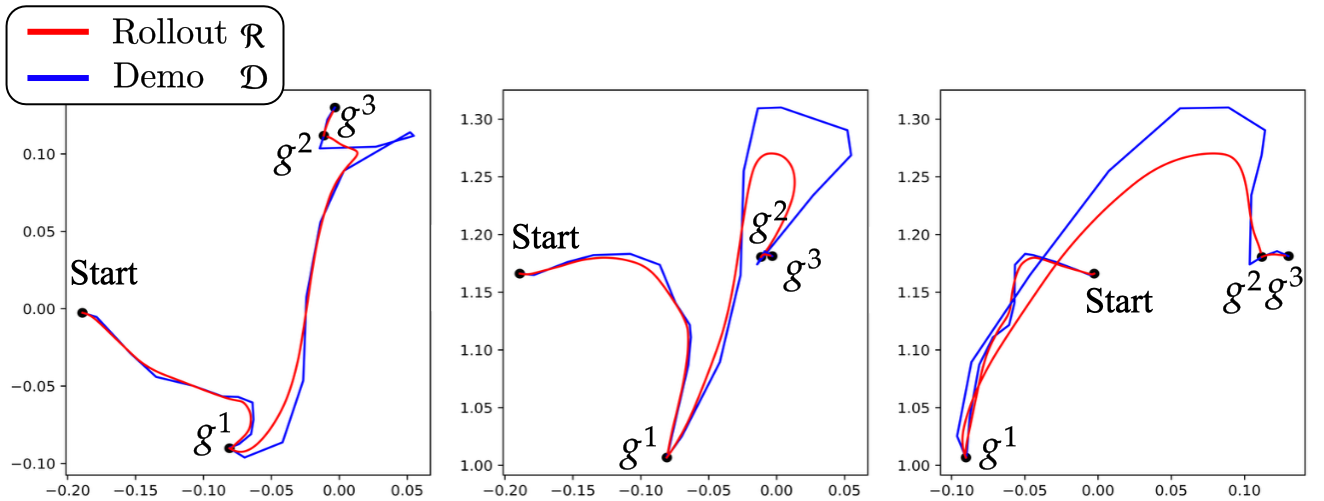}
        \caption{{\Ours} }
        \label{fig:Ours2d}
    \end{subfigure}
    \caption{2D projections (from left to right: xy, xz, yz) of policy rollouts (red) learned from demo (blue) in {\bf Ketchup} task in a \textit{deterministic} environment.}
    \label{fig:exp-robosuite}
\end{figure}

\subsection{Experimental setups}
\noindent 
We evaluate our work in Robosuite~\cite{zhu2020robosuite} with tasks defined in public benchmark LIBERO~\cite{liu2023liberobenchmarkingknowledgetransfer} and robomimic~\cite{mandlekar2021matterslearningofflinehuman} (see Fig.~\ref{fig:exp-robosuite-tasks}).
We selected the following pick-and-place tasks from the benchmark:
\begin{enumerate}
\item Ketchup: grab a ketchup bottle and place it in a drawer.
\item Square: pick up a square nut and fit it on a square rod.
\item Wine: grab a wine bottle and set it on a rack.
\item Bowl: lift a small bowl and place it on a cabinet.
\item Cheese: pick up a bar of cheese and put it in a bowl.
\end{enumerate}

\noindent 
Each demonstration contains approximately 100 to 250 data points.
We use our method described in Sec.~\ref{method:segmentation} to divide the demonstration and select waypoints with an error threshold of $\AWEthreshold = 0.01$ for each segment.
This yields 10 to 25 waypoints from the demonstration.
See Fig.~\ref{fig:exp-robosuite-tasks} for more information on the tasks we evaluate on.

We define the state of the system at the end-effector position and the gripper state. The control commands are generated through standard inverse kinematics.
For each task, all policies are trained from {\bfseries one demonstration} over $10,000$ epochs. We evaluate both the overall task completion and the success of each individual subgoal. That is, we assess subgoals independently—if subgoal 1 fails (due to a timeout), the environment is reset to the state of a successful subgoal 1, and the simulation continues with subgoal 2. Evidently, if the model completes the task after a reset to a later subgoal, it is not considered a successful attempt. All experiments are done with a maximum horizon of 1000 actions per subgoal.

During execution,
the learned dynamical system outputs the linear velocity of the end-effector based on the current end-effector position. 
The angular movement is calculated separately using simple spherical linear interpolation (SLERP), as most of the manipulation tasks do not involve complex movement in the orientations. 
As long as the gripper reaches the subgoal with the correct orientation, the speed and timing of orientation adjustments do not affect task success. 
At each time-step, $\controller$ evaluates whether the current sub-goal was achieved, based on a distance threshold $\thresholdK = 0.008$m, which is precise enough for pick-and-place tasks. 

\begin{table}[t]
\centering
\setlength{\tabcolsep}{3pt}
\caption{Success rate (\%) for behavior cloning benchmark and our work in \textit{noisy} and \textit{perturbed \& noisy} environments across various tasks.}
\begin{tabular}{@{}lcccccccc@{}}
\toprule
\multirow{2}{*}{\textbf{Task}} & \multirow{2}{*}{\textbf{Subgoal}} 
& \multicolumn{2}{c}{\textbf{Noisy}} & \multicolumn{2}{c}{\textbf{Perturbed \& Noisy}} \\
\cmidrule(lr){3-4} \cmidrule(lr){5-6}
  & & \BCSeg & \Ours & \BCSeg & \Ours \\ \midrule
Ketchup & 1       & $96.7 \pm 5.8$   & $\mathbf{100.0 \pm 0.0}$  & $73.3  \pm 5.8$   & $\mathbf{93.3  \pm 5.8}$ \\
        & 2       & $\mathbf{100.0 \pm 0.0}$  & $\mathbf{100.0 \pm 0.0}$  & $\mathbf{96.7  \pm 5.8}$   & $\mathbf{96.7  \pm 5.8}$ \\
        & 3       & $\mathbf{100.0 \pm 0.0}$  & $\mathbf{100.0 \pm 0.0}$  & $96.7  \pm 5.8$   & $\mathbf{100.0 \pm 0.0}$ \\
        & total   & $96.7  \pm 5.8$   & $\mathbf{100.0 \pm 0.0}$  & $60.0  \pm 10.0$  & $\mathbf{80.0  \pm 0.0}$ \\ \midrule
Square  & 1       & $\mathbf{100.0 \pm 0.0}$  & $\mathbf{100.0 \pm 0.0}$  & $\mathbf{70.0 \pm 0.0}$    & $\mathbf{70.0 \pm 0.0}$ \\
        & 2       & $\mathbf{100.0 \pm 0.0}$  & $\mathbf{100.0 \pm 0.0}$  & $66.7 \pm 5.8$    & $\mathbf{70.0 \pm 0.0}$ \\
        & 3       & $\mathbf{100.0 \pm 0.0}$  & $\mathbf{100.0 \pm 0.0}$  & $90.0 \pm 17.3$   & $\mathbf{100.0 \pm 0.0}$ \\
        & total   & $\mathbf{56.7 \pm 15.3}$  & $\mathbf{56.7 \pm 11.5}$  & $53.3 \pm 11.5$   & $\mathbf{70.0 \pm 0.0}$ \\ \midrule
Wine    & 1       & $\mathbf{100.0 \pm 0.0}$  & $\mathbf{100.0 \pm 0.0}$  & $\mathbf{90.0 \pm 0.0}$    & $\mathbf{90.0 \pm 0.0}$ \\
        & 2       & $\mathbf{100.0 \pm 0.0}$  & $\mathbf{100.0 \pm 0.0}$  & $30.0 \pm 30.0$   & $\mathbf{70.0 \pm 0.0}$ \\
        & 3       & $\mathbf{100.0 \pm 0.0}$  & $\mathbf{100.0 \pm 0.0}$  & $30.0 \pm 30.0$   & $\mathbf{80.0 \pm 17.3}$ \\
        & total   & $\mathbf{63.3 \pm 15.3}$  & $\mathbf{63.3 \pm 15.3}$  & $20.0 \pm 17.3$   & $\mathbf{50.0 \pm 10.0}$ \\ \midrule
Bowl    & 1       & $90.0 \pm 17.3$   & $\mathbf{100.0 \pm 0.0}$  & $56.7 \pm 5.8$    & $\mathbf{93.3 \pm 11.5}$ \\
        & 2       & $\mathbf{100.0 \pm 0.0}$  & $\mathbf{100.0 \pm 0.0}$  & $90.0 \pm 10.0$   & $\mathbf{100.0 \pm 0.0}$ \\
        & 3       & $\mathbf{100.0 \pm 0.0}$  & $\mathbf{100.0 \pm 0.0}$  & $93.3 \pm 11.5$   & $\mathbf{97.6 \pm 5.8}$ \\
        & total   & $83.3 \pm 20.8$    & $\mathbf{97.6 \pm 5.8}$   & $46.7 \pm 15.3$   & $\mathbf{70.0 \pm 0.0}$ \\ \midrule
Cheese  & 1       & $80.0 \pm 26.5$    & $\mathbf{100.0 \pm 0.0}$  & $33.3 \pm 32.1$   & $\mathbf{60.0 \pm 26.5}$ \\
        & 2       & $100.0 \pm 0.0$    & $\mathbf{100.0 \pm 0.0}$  & $53.3 \pm 20.8$   & $\mathbf{60.0 \pm 0.0}$ \\
        & 3       & $100.0 \pm 0.0$    & $\mathbf{100.0 \pm 0.0}$  & $90.0 \pm 17.3$   & $\mathbf{100.0 \pm 0.0}$ \\
        & total   & $80.0 \pm 26.5$    & $\mathbf{93.3 \pm 11.5}$  & $16.7 \pm 11.5$   & $\mathbf{50.0 \pm 10.0}$ \\ \bottomrule
\end{tabular}
\label{table:combined-success-rates}
\end{table}

\subsection{Results in Deterministic Environments}

\noindent 
We begin by evaluating all baseline models and our approach in deterministic environments. Each model is trained using 10 different random seeds, and the simulation is ran once to assess success or failure.  Our results demonstrate {\bf perfect success rates} (100\%) across training seeds for both {\Ours} and {\BCSeg}, while {\BC} and {\SNDS}, which lack segmentation, exhibit {\bf near-zero success rates}—even when success is evaluated at the subgoal level. Figure~\ref{fig:exp-robosuite} compares the rollouts of each model with the demonstration, clearly highlighting the poor performance of non-segmented models in contrast to their segmented counterparts. Since \BCSeg ~has a similar performance with \Ours ~in this experiment, only one plot is shown here.


This behaviour is expected since dividing the task into segments and then sampling the waypoints drastically simplifies the complexity of the motion and learning becomes easier. Also, those methods learn to be conditioned on achieving a specific sub-goal.
Methods that take the whole demonstration as an input are unlikely to learn well with a single demonstration.

   
\subsection{Rollouts with Noise and Perturbations}
\begin{figure}[t]
    \centering
    \begin{subfigure}[b]{\linewidth}
        \centering
        \includegraphics[width=\textwidth]{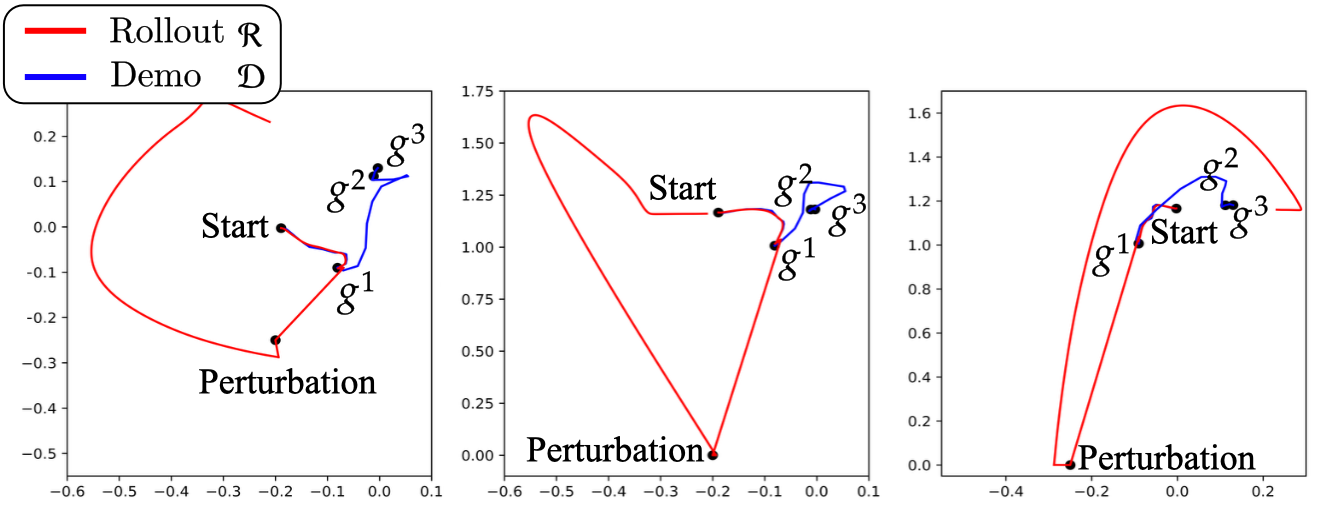}            
        \caption{\BCSeg}
        \label{fig:perturbation-nn}
    \end{subfigure}
    \hfill \\
    
    \begin{subfigure}[b]{\linewidth}
        \centering
        \includegraphics[width=\textwidth]{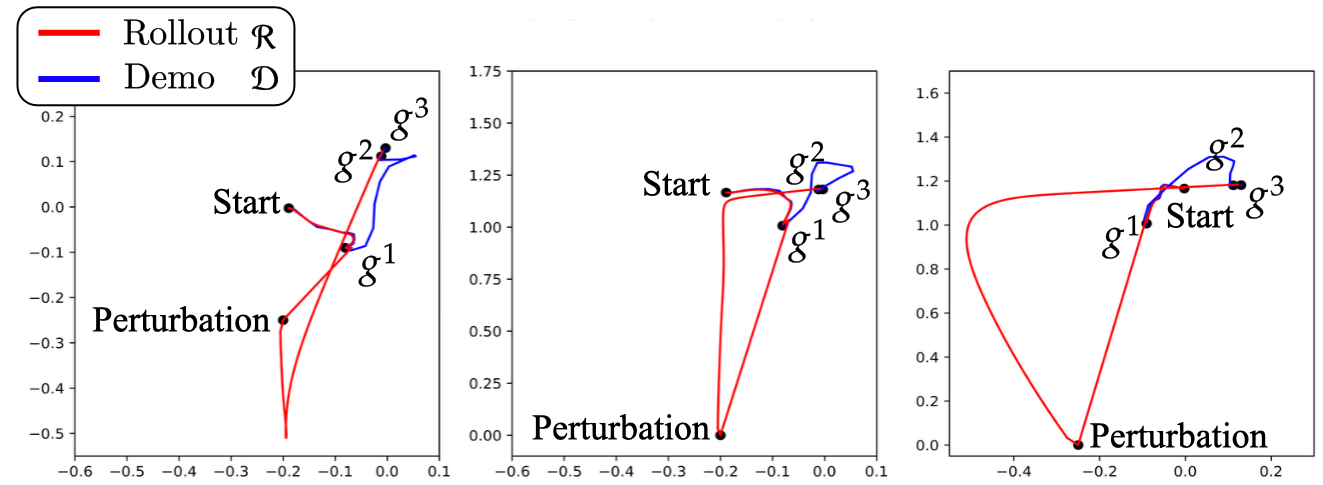}
        \caption{\Ours}
        \label{fig:perturbation-snds}
    \end{subfigure}
    \caption{2D projections (from left to right: xy, xz, yz) of policy rollouts (red) learned from demo (blue) in {\bf Ketchup} task with perturbations injected.}
    \label{fig:exp-robosuite-perturbation}
\end{figure}

\noindent
To further evaluate how a learned policy performs in a realistic setting, we add noise and perturbations during rollouts.
Since $\BC$ and $\SNDS$ have not achieved any tasks in a deterministic setting, we only compare $\BCSeg$ and $\Ours$ in this section. 

We add Gaussian noise with a standard deviation of $0.01$ to the end-effector position feedback. This is comparable to the noise level of a real robotic system.
In a real-world application, the robot might experience unexpected disturbances (e.g., from inexperienced users). 
For this, we also investigate the effect of perturbations while executing the policy. 
We train different policies using 3 random seeds and evaluate each model 10 times. We repeat the same experiments and the average success rates across all 5 tasks are reported in Table~\ref{table:combined-success-rates}.

For both scenarios, we can see that $\Ours$ consistently performs better than $\BCSeg$.
This is due to our choice of dynamical systems that enforce the movement toward each sub-goal. 
Therefore, it is robust under noise and disturbances.

To visualize the performance of the rollouts, we plot the demonstration and the rollout in Fig.~\ref{fig:exp-robosuite-perturbation}. 
We can see that the rollout generated from $\BCSeg$ (Fig.~\ref{fig:perturbation-nn}) diverges from the trajectory and is unable to recover. 
Note that, the rollouts are generated using models trained from data using our segmentation and waypoint methods described in Sec.~\ref{method:segmentation}, and each segment is a relatively simple task.
However, this phenomenon is expected since the perturbation pushes the robot to a region that is not covered by the demonstration. 
Similar to most standard data-driven algorithms, the model does not respond well to out-of-distribution scenarios. 

In contrast, as seen from Fig.~\ref{fig:perturbation-snds}, $\Ours$ is robust to perturbations and re-plans as needed. 
Even with perturbations, the constraints imposed by the Lyapunov condition can guide the robot back to the next subgoal.
While most work in literature requires a large dataset to train a policy, our work only requires a single demonstration.

\begin{figure}[t]
      \centering
      \includegraphics[width=0.43
\textwidth]{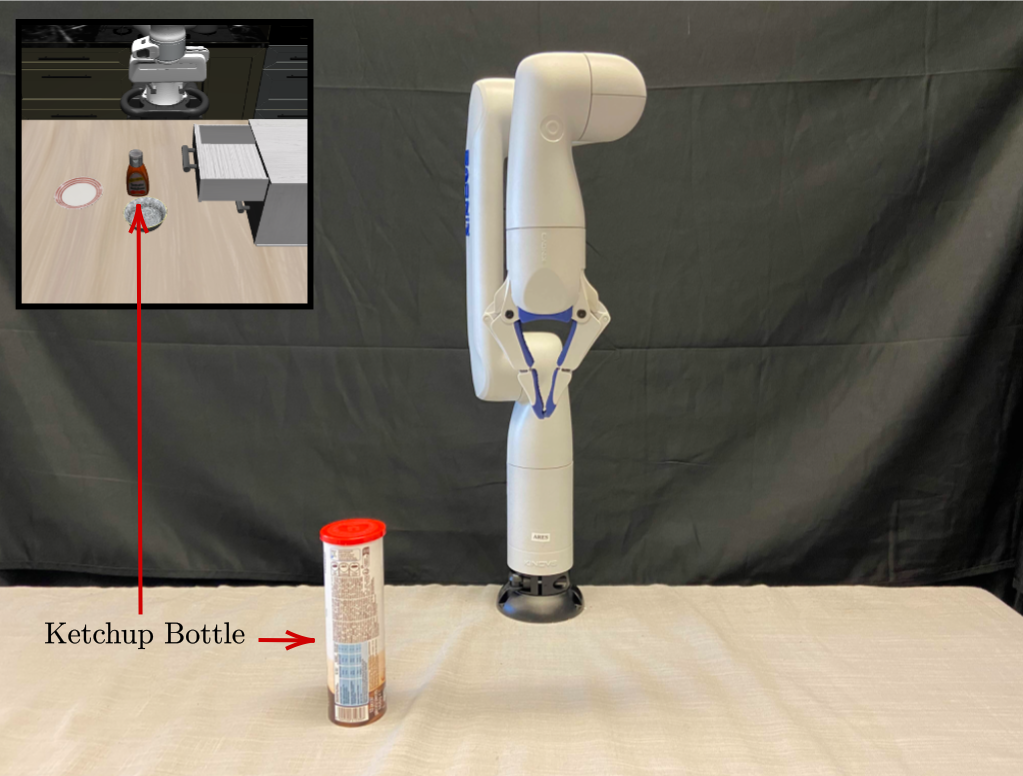}
      \caption{Visualization of tasks on robotic hardware.}
      \label{fig:exp-robot}
\end{figure}

\subsection{Zero-shot sim-to-real}
\noindent
The dynamical systems were trained with {\em one demonstration} in simulation, without further data augmentation. 
We deploy the dynamical systems learned from the simulation on robotic hardware (see Fig.~\ref{fig:exp-robot}). 
Note that, we did not observe any {\em sim-to-real gap}.
This is because the Lyapunov condition ensures that the motion is constrained toward sub-goals, diminishing the likelihood of divergence. 
   

%% file: tex/conclusion.tex
\noindent 
In this paper, we build upon prior work on learning long-horizon manipulation tasks and stable learning with dynamical systems, with the goal of improving task success rates of execution while minimizing the required training data. We introduce a novel approach that (1) decomposes long-horizon demonstrations into segments defined by waypoints and subgoals, and (2) learns globally stable dynamical system policies that drive the robot toward each subgoal, even in the presence of sensory noise and stochastic disturbances. The proposed method is validated both in simulation and on robotic hardware, demonstrating seamless transfer from simulation to real-world implementation.

In this work, we demonstrate a proof-of-concept using a sequence of stable dynamical systems for long-horizon manipulation tasks. 
Our future work will investigate various automatic segmentation methods with visual feedback~\cite{zhang2024universal,koukisa} and make our work more generalizable to unseen scenarios. We will also apply this learning regime to a more complex state representation.
